\documentclass[runningheads]{llncs}

\usepackage[T1]{fontenc}
\usepackage{graphicx}
\usepackage[dvipsnames]{xcolor}
\usepackage{hyperref}
\usepackage{amssymb}
\usepackage{todonotes}

\definecolor{scopecolor}{HTML}{00aff0}

\newcommand{\risk}[1]{\textcolor{Bittersweet}{\textbf{#1}}}
\newcommand{\scope}[1]{\textcolor{scopecolor}{\textbf{#1}}}
\newcommand{\benefit}[1]{\textcolor{DarkOrchid}{\textbf{#1}}}
\newcommand{\safeguard}[1]{\textcolor{SeaGreen}{\textbf{#1}}}
\newcommand{\software}[1]{\textcolor{black}{\textbf{#1}}}

\usepackage{biblatex} 
\begin{document}

\title{Concepts for Securing Agentic AI Coding \texorpdfstring{\newline}{} and the Terok Environment}
\titlerunning{Securing Agentic AI Coding and the Terok Environment}

\author{Jiří Vyskočil\inst{1,2}\orcidID{0000-0001-8822-0929} \and
Franz Pöschel \inst{1,2}\orcidID{0000-0001-7042-5088} \and
Andreas Knüpfer \inst{1,2}\orcidID{0000-0003-3591-397X}}

\authorrunning{Jiří Vyskočil, Franz Pöschel, Andreas Knüpfer}

\institute{Center for Advanced Systems Understanding (CASUS)\\in Görlitz/Saxony/Germany
\url{https://www.casus.science} 
\and
Helmholtz-Zentrum Dresden-Rossendorf (HZDR)\\in Dresden/Saxony/Germany
\url{https://www.hzdr.de}
}

\hypersetup{%
  pdftitle={Concepts for Securing Agentic AI Coding and the Terok Environment},
  pdfauthor={Jiří Vyskočil, Franz Pöschel, Andreas Knüpfer},
  pdfsubject={IT security concepts and implementation for agentic AI coding and the Terok environment},
}

\maketitle

\begin{abstract}
Agentic AI is a fascinating new tool for software development. It is a huge step forward compared to "conventional" AI assisted coding, which in turn was a considerable breakthrough earlier. AI support through LLMs is a young and very fast-moving field. The "conventional" (non-agentic) flavor became useful and productive in early 2025 (around 18 months ago) and the agentic flavor followed in fall 2025 (approximately 9 months ago).
Besides all its benefits and potential, it also carries some fundamental risks for IT security. And the agentic approach added very severe risks while making others much more dangerous.

With all the motivation to explore this fascinating new tool we should not ignore the risks but actively address them. We present (I) an assessment of the IT security risks, (II) a concept for mitigating them without breaking its benefits, and (III) an overview about an implementation of our concept. In this very dynamic field this is likely not the final and once-and-for-all answer to the identified issues but still a substantial step forward in responsible usage of Agentic AI for software development. It should also be a contribution to the community to allow early and eager evaluation of the potential of agentic AI for software development without actually suffering from its implied IT security risks.
\end{abstract}

\section{Introduction}

AI as an assistant for coding became feasible and useful with the recent breakthrough in Large Language Models (LLM) which work for programming languages as well as for natural languages.
This caused many debates already, in the same way as it did for natural language processing. There are strong opinions how AI assisted coding should or should not be handled. They range from the belief that \emph{AI will replace all programmers very soon} to the conviction that \emph{no programmer must use any AI}. The authors believe both extremes are wrong and personally don't like to give up this tool anymore.
At the same time, we want to examine the implications of these new tools and start working to mitigate some risks we identified. In this paper we focus on the application of Agentic AI for coding and look at severe IT security issues emerging from it. We present a proof-of-concept solution and want to participate in the necessary discussion how to shape a responsible manner for using Agentic AI in software development.

\subsection{What is AI Coding?}

The more general term “AI coding” (without an “agent”) refers to any general form of coding with the assistance of artificial intelligence; yet at the time of writing and for all practical purposes it means with support by a Large Language Model (LLM).
It became also known as ``Vibe Coding'', a term coined in February 2025 with a negative connotation suggesting that it is used (only) by programmers who would not be able to program without AI support. In any case, this marks the time when AI assisted coding became usable and useful enough for novice programmers to benefit from it.

It started with general chatbot interfaces where a programmer could ask an LLM for source code as part of a normal conversation. Programming IDEs added more convenient interfaces which made it easy to present existing code to the LLM and insert generated code to source code files. All additional functionalities of IDEs like formatting, linting, syntax checks, compiling, debugging, testing etc. were more or less the conventional (non-AI) tools as had existed before.

\subsection{What is Agentic AI Coding?}

Agentic AI coding represents the next major step after the AI chatbot mode. It provides an additional way of interaction where the LLM can "use a tool" as an answer. This is an action chosen by the LLM to be executed in the user's environment such as reading something on the user's local file system, writing or modifying files, or executing an arbitrary command. Furthermore, the LLM receives the output of the tool calls and can react on it.

As a concept, the term ``Agentic AI'' is much older than ``Vibe Coding'' but only became usable for software development purposes after "Vibe Coding".

\section{We Need Unrestricted Agentic AI for Coding!}
\label{sect:benefits}

The “agentic” aspect to AI programming assistants is not just a slight improvement but a fundamental change of paradigm when it can run in an unrestricted way. This means not restricted by continuous user supervision, no interactive step-by-step approval by the user but allowing long phases of working independently from the human user. 
This is for several reasons:



First, intelligent automation spans a wider scope of interactions. Before, the LLM produced some code, the IDE inserted it at the correct position, and then the user had to continue with the next step like compiling or testing. 
Now, the agent directly calls the needed commands to compile the code, execute the results, or run automated tests. This eliminates one extra user interaction (prompt) and is at the same time the core which makes the LLM have its own agency (\benefit{AGENCY BENEFIT}).

The second reason is how the conversation with the LLM continues. Since the LLM will receive the output of its “tool calls” (the commands it executed in the user's environment) it can react to it. Instead of waiting for the next user prompt, it proceeds autonomously (\benefit{AUTONOMOUSLY PROCEED BENEFIT}). For example, it reacts to the output of the compiler. If the compiler reports a successful build, then the agent goes to the next step and executes the new binary. In case the compiler (or any other step) reports an error, the agent will try to fix it and then try again. The same is true for configuration steps or test runs for code.
Together with the previous benefit, this will not only save one user prompt but many in a row and no longer requires the user to drive the conversation with the LLM step by step.

Furthermore, when the agent proceeds autonomously like this, it will try to continue until its task is accomplished. If there is an error left, it will try to correct it before coming back to the user. This is a self-correcting manner and the user will observe much more adequate answers (\benefit{SELF-CORRECTING BENEFIT}). 
This demonstrates that the accuracy of an LLM's responses may be increased not only by improving the model itself, but also by applying the same unaltered model within a more powerful infrastructure.
%

If information is missing -- it may be an unknown answer to an open problem (like "How to call a given command line tool to solve one subtask?") or simply required data (like "What is the peak floating-point performance with vector instruction on the given CPU?") -- then the agent can query it. Either from the environment or from sources on the internet -- for the two examples here, both ways would be possible (\benefit{AUTONOMOUS QUERY BENEFIT}).

Finally, the agent may install dependencies, meaning it can modify its own development environment (\benefit{MODIFY ENVIRONMENT BENEFIT}). This is one more step it can solve itself without asking the user to do it.

All those benefits provide a faster and more efficient work style especially w.r.t. the required user interactions. And it allows a single user to employ many agents to work basically autonomously on different sub-tasks. All benefits combined make it a very promising tool and we should definitely explore it further.

Yet, these benefits are in diametrical conflict with Section~\ref{sect:dangerous} and the risks detailed therein. Furthermore, the listed benefits can also turn into risks when they work against the user. 
Therefore, we will call them "\textbf{BENEFIT/RISK}".

Now, can we find a way to profit from the benefits and at the same time avoid or at least control and mitigate the risks?
The authors are convinced that we (the scientific and software developer communities) need to explore this fascinating new tool of Agentic AI for coding. We should not leave it to the carefree and naive to explore first, and they should certainly not set the standards for all of us. We believe that we can provide a sufficiently safe and secure environment for the purpose of agentic software development, although this is not yet feasible for general AI agents and it is unclear if this will ever be possible. Our goals include protecting the development environment, protecting secrets and credentials, protecting source code repositories, and adding convenience. In Section~\ref{sect:concepts} we will present and discuss our approach and our implementation.

\section{Why is Agentic Coding Dangerous?}
\label{sect:dangerous}

We need to think about security implications of agentic AI for software development, because there are consequences beyond the known risk during the software development and lifecycle phases. There are old risks which become more severe when AI assistants are involved and there are entirely new kinds of risks.


\subsection{Old But More Dangerous Risks}

Among the existing risks there are some which become more dangerous with agentic AI.
%
%
Software supply chain attacks count among them. There are two flavors. The first is the generated software where malicious AI tools might try to sneak in malicious code unbeknown to the user (\risk{MALICIOUS CODE GENERATION RISK}).
This also overlaps with the following groups of risks. It definitely becomes more dangerous in combination with agentic AI because such malicious code may be generated and executed almost simultaneously by the coding agent.
The second flavor is software supply chain attacks in software for agentic AI (\risk{SOFTWARE SUPPLY CHAIN RISK}). Software supply chain attack means that malicious code is not inserted in the target software project but in one of the dependencies (libraries, modules, ...). It is particularly devious when previous versions of a dependency were trustworthy but became malicious during a minor version upgrade.
This type of attack is noteworthy because there are several such attacks focusing on agentic AI in particular%
\footnote{Noteworthy ones are among many others the attack on the Axios HTTP client from March 2026 (\url{https://socket.dev/blog/axios-npm-package-compromised}) and on the MistralAI client from May 2026 (\url{https://docs.mistral.ai/resources/security-advisories})}.

\subsection{New Risks}

There are the \textit{early AI risks} from when the use of AI assistants (without the "agent") for programming was starting to become commonplace.
%
%
The first group of early AI risks concerns incorrect answers by the agent. 
This may be by mistake because the LLM doesn't know better (\risk{RISK OF MISTAKES}). This also touches so-called "hallucinations" where the LLM fabricates answers instead of admitting that it doesn't know (\risk{RISK OF HALLUCINATION}).

It may also be due to "prompt injections" where someone tricks the LLM into accepting alternative instructions (\risk{PROMPT INJECTION RISK}). Finally, the agent / the LLM may act on evil intentions that no one instructed it to (\risk{EVIL AGENT RISK}) \footnote{See the "agentic misalignment" where AI acts in malicious ways to avoid its own shutdown \url{https://www.anthropic.com/research/teaching-claude-why}} \cite{macdiarmid2025natural}.

All four flavors become more dangerous due to the agent's ability to act on its own in the user's development environment. And it adds to the risk that the agent inserts such results in the software it helps generate.

The scope and power of AI systems come with an inherent risk of undesired behaviors, including for example wrong answers or an agent's apparent hesitance from admitting that it does not know an answer. As eliminating such behaviors is a stretch goal at best, any reasonable real-world use must accept their possibility and explore effective countermeasures toward their mitigation.



One aspect of this is the dependency on commercial providers (\risk{VENDOR DEPENDENCY RISK}), especially "Big Tech" from the U.S. who may discontinue services for end users or (drastically) increase pricing (as already happened multiple times for the \emph{API access} which is required for “agentic mode”).
The IT security aspect is that the vendor can read all your prompts and files and can execute arbitrary commands in your local environment (\risk{VENDOR TRUST RISK}).

Both of these aspects cannot be avoided when using commercial services. Yet, there is the alternative of self-hosted models. A number of open models (Open Source or Open Weights models) are available and show promising abilities. They are certainly a good alternative for many kinds of programming tasks. However, many of the (currently) leading models are not released for self-hosting. Apart from the individual decision which model to use for a given programming job, we should include both kinds of models in our ongoing exploration of AI-assisted software development. In any case, we need to preserve the ability to self-host LLMs.

There is also the more general dependency of human programmers on AI tools. Would humans forget how to program after relying on AI tools for too long? The danger of "skill atrophy" for humans does exist, however the authors believe it is out of reach of a technical solution anyway. We will not discuss it further here. (\risk{SKILL ATROPHY RISK})

\subsection{Brand New Risks}

The completely new risks originate from the agent's ability to act inside the user environment. This “agency” is exactly what makes it more valuable but we do not (yet) know how to control it and prevent unintended consequences.

To shine some light on it, we compare agentic AI for programming to general AI agents without any restricted purpose. It may be used for general tasks such as answering emails, organizing appointments, booking restaurants, online shopping, ... all the way to difficult math problems and science. It can execute arbitrary commands or access remote resources. For this purpose, the agent is given necessary credentials on purpose, such as the email and calendar account passwords or credit card numbers. With this the general AI agent acts on behalf of the user and performs either desired or undesired tasks.

A well known example is OpenClaw (formerly ``MoltBot'', formerly ``Clawdbot''), started by a single developer from Austria \cite{openclaw}, which has been utilized in a naive and carefree way by users unaware of potential consequences.

With OpenClaw and similar, the agent can act like a loyal servant or a virus or backdoor. Both is technically the same, only the (assumed) intentions differ. In combination with the \risk{RISK OF MISTAKES}, the \risk{EVIL AGENT RISK}, the \risk{VENDOR TRUST RISK}, and the \risk{PROMPT INJECTION RISK} this produces the risk of a behavior of a virus or backdoor (\risk{VIRUS OR BACKDOOR RISK}). As part of this it can find and exfiltrate sensitive data (\risk{EXFILTRATE DATA RISK}).

The \emph{exfiltrate data risk} of autonomous agents has consequences undermining traditional assumptions for the design of authentication systems. As a particularly puzzling instance of this risk, an adversarial agent may run at danger of exfiltrating to a third party the very key\slash token that it needs for authenticating with the AI\slash LLM service provider%
\footnote{The software supply chain attack on LiteLLM from March 2026 tried this, see \url{https://futuresearch.ai/blog/litellm-pypi-supply-chain-attack/}}.
Hence, the key must be hidden from precisely that software component that is supposed to use it. This implies that any sensitive information of the kind must not be available anywhere within an agent's boundary of free movement, e.g.\ a container or VM.

Transferring these considerations back to the restricted purpose of using AI agents for software development, we observe essentially the same situation: If an AI coding agent is used in the user's environment on a personal laptop, then it has the same access permission as the user. Thus, it could also access secrets and login credentials and use them. In summary, AI agents for coding share the same potential for unintended actions as have been observed in OpenClaw! And thus all of its risks apply, too.

%
%
%

\section{Guardrails for AI Coding Assistants and Related Work}



There are two types of conventional (non-AI) measures to guardrail AI behavior and results as well as two AI-driven approaches. Here, "guardrail" means prevent unintended consequences in the broadest sense. It is a vague term because it should comprise all the risks and benefits for all the scopes detailed in Sections \ref{sect:benefits}, \ref{sect:dangerous}, and \ref{sect:concepts}.

The first guardrail attempt is step by step confirmation for each tool command by the agent. The first agentic clients used it and it is certainly effective: if the user checks and approves each command he or she can prevent unintended actions. However, it requires constant attention by the user and more or less reduces the interaction to the older "chatbot mode". This would erase all advantages of the agentic mode. Furthermore, from a usability standpoint, most human developers would surrender to "allow all" sooner or later or even look for ways to bypass this guardrail.

The second conventional approach is restrictions of allowed commands. This defines allow-list or deny-lists of commands and options to commands and then disallows certain tool calls by the agent. This is not secure because agents find clever ways to work around such rules -- it is not important if they do so with good intentions to achieve a goal given by user or with malicious intent. There are reports that AI agents evade a ban of the git command by using the git API in a Python script. This trick could be transferred to almost all commands.
In the end, this strategy fails due to the lack of bulletproof rules.

The first AI based guardrail is to instruct the agent/LLM not to do certain actions. This can be part of a system prompt or an explicit command by the user. From many personal experiences and reports we know that agents sometimes obey and sometimes disobey. Often, agents will even apologize afterwards, which shows they didn't forget such an instruction. Still, sometimes they disobey it again shortly afterwards. So this approach is not sufficiently safe and secure. The current state of the art in training LLMs has no solution how obedience to such instructions can be guaranteed.

The AI based guardrails can be extended by adding a control AI to watch over the actions by the coding AI agent. This can be effective and allows the controlling agent to be instructed with a separate set of commands toward another goal. Still, the fundamental loophole persists: If one agent can be tricked into malicious actions then two agents might as well.


Even though Agentic AI coding is a very young topic it found attention by the academic community already.



Kozak et. al. conducted one of the first systematic safety evaluations of autonomous coding agents, analyzing over 12,000 actions across five state-of-the-art models on 93 real-world software setup tasks and developed a high-precision detection system that identified four major vulnerability categories \cite{kozak2025developer}. As mitigations they suggest security reminders and feedback mechanisms to the agents to make them pay more attention to security aspects.


Mukherjee et.al. analyzed the same problem and observed that security flaws accumulate over successive refinements (development steps) by AI. Their attempt at a solution consists of a multi-agent framework for security checks and control \cite{mukherjee2026multi}. 
This means, they propose to use more AI agents to check on the coding AI agents. On the one hand this can be promising because dedicated AI agents are able to help and it would not cost human effort. On the other hand it will increase the number of agents one has to trust and to supervise. One could argue that it will merely shift the problem but not solve it.


Bowers et.al. find that multi-agent systems for AI coding are highly vulnerable to attacks and show that some multi-agent architectures are more robust and others are more susceptible \cite{bowers2025analyzing}. They conclude that humans are required as part of the review process for AI generated code -- "human-in-the-loop" is required and should not be replaced by a fully autonomous setup. We agree to this statement and see it as an essential building block to our approach. 
They also comment on the increased attack surface when AI agents control AI coders ("the
security analysis agent is vulnerable ...").



In more general terms, not restricted to the coding tasks but for general AI agents and the LLMs behind them, Zheng et.al. show how LLMs are prone to system prompt extraction attacks. This means, the LLMs without the additional components in the agentic coding setup. They underline that "agentic interaction fundamentally expands the LLM attack surface" \cite{zheng2026just}. For our approach presented here this encourages our aspect of protecting the LLM scope.


The study by Siddiq et.al. presents a large-scale empirical analysis of agent-authored PRs (over 33,000 curated PRs) and analyzes prevalence, acceptance outcomes, and review latency across autonomous agents, programming ecosystems, and types of code changes \cite{siddiq2026security}.
They find that human effort for proper code reviews grows, especially for thorough security checks. So the human effort in the complete software development process may not even decrease when AI agents support humans with generating code.


Other literature is looking at general AI agents (without the restriction to software development tasks) and their security implications.



Maybe the most prominent one is OpenClaw (also see above). The study by Chen et.al. states the "heightened safety and security concerns". It identifies six risk dimensions which are not basic software security risks but rather different ways of wrong/unexpected/disloyal behavior by the agents \cite{chen2026trajectory}.
Interesting insights are that more failures arise under under-specified intent, open-ended goals, or even benign-seeming jailbreak prompts. Also minor misinterpretations may escalate into "higher-impact tool actions" and "propagates unsupported assumption". Those effects could certainly transfer to AI agents for coding.

Comparable efforts to sandbox coding agents differ mainly in their confinement method:
Docker Sandboxes~\cite{dockersandboxes} (micro-VM), microsandbox~\cite{microsandbox}
(libkrun micro-VM), nono~\cite{nono} (OS-level Landlock/Seatbelt), and NVIDIA
OpenShell~\cite{openshell} (container or micro-VM). The most prominent of these, Docker
Sandboxes, is closed-source; Terok instead provides a fully open-source 
environment built natively on the rootless Podman stack, and is vendor-independent across
  commercial, self-hosted, and national or academic LLM services.

\section{Concepts for Safeguarding Unrestricted Agentic AI and the Terok Environment}
\label{sect:concepts}

\noindent\textbf{Trust model.} We treat the coding agent --- the LLM, any instructions injected into 
it through the data it reads, and the code it produces --- as untrusted. We trust the host, 
its operating-system kernel, and the isolation that \software{Podman} (optionally via \software{krun}) 
enforces, together with the safeguard components running outside the container. Container or micro-VM 
escape, a compromised host, and the trustworthiness of the vendor model itself (\risk{VENDOR TRUST RISK}) are
therefore out of scope here.
  
We propose separating the security concerns into five scopes and add four main security safeguards.

\subsection{Scopes}

The most important scope that should be protected is the developer's local environment, that is the personal laptop or workstation or development server (\scope{LOCAL SCOPE}). The agent should not get full control over the local environment, because there will be personal data as well as unrelated code, data, workflows etc. which are not needed for the agent's tasks in the given project. First and foremost, the local environment is separated from the (\scope{CONTAINER SCOPE}) inside of which the agent can run unrestricted.

The next scope is the git forge repository for the given project, that is the central GitHub or GitLab or similar repository for collaborative development together with other humans or agents (\scope{GIT FORGE SCOPE}). Besides the primary code repository this also involves team management (members and roles), CI, releases etc.

The third scope is that of the LLM provider (\scope{LLM PROVIDER SCOPE}) which can either be a commercial vendor, non-commercial providers such as e.g.\ national academic institutions, or your own institution with self-hosted LLMs. Those may need different levels of protection. The LLM provider is a separate scope because the protection goes two ways. On the one hand, one needs to trust the LLM provider to some extent, still it should not become a part of the local scope. On the other hand, the LLM provider should be protected against malicious code which tries to employ it for someone else's purposes or to steal the access credentials for it.

Finally, there are other points of interaction like data sources, many types of sources for software components to be installed (pip, npm, dockerhub, etc. ), or search engines or general web resources to query. They are summarized under “anything else” (\scope{REST OF THE WORLD SCOPE}).

\subsection{Safeguards}

The primary safeguard%
\footnote{This first safeguard was the author's first step towards the presented solution and the initial assumption was that it would be sufficient. Then it would probably not justify a special software solution nor this paper, yet we quickly became aware of more security concerns which immediately follow.}
is an isolated execution environment (\safeguard{CONTAINER ENVIRONMENT SAFEGUARD}) where the agent can operate in an unrestricted manner (according to the above definition without close supervision). This can be a software container or a virtual machine. By default we use the more lightweight container runtime, and additionally support \software{krun}-based micro-VMs for stronger isolation.

The second safeguard which is also second in importance is about communication with the outside world, especially outbound requests (\safeguard{EGRESS FIREWALL SAFEGUARD}). This will allow precise control of external communication partners. The protection goes both ways. It can protect external servers from malicious access by the agent inside the container and also protect the agent from accidentally downloading questionable code or data from the internet. To be effective it should default to "deny" and provide an easy allow/deny management for the user.

The third safeguard is a git gateway service (\safeguard{GIT GATEWAY SAFEGUARD}) which acts as a delayed proxy to the central git forge repository (see \scope{GIT FORGE SCOPE} above). It is a local mirror of the remote repository on the host system, acting as a bridge between the host and the container. The git gate uses the decentralized design of the git version control system to replace the upstream git remote, this way becoming visible to the agent in lieu of the production repository. So, the agent can do commits, branches, merging, ... without the user's supervision. 
However, any of its changes, including potential wrongdoing, are intercepted by the git gate, while hiding from the agent the production version of the repository where unwanted actions may have hard to solve consequences. Instead, the user should check the results in the local gateway repository and forward it to the public production repository if approved. This is still an advantage for the user because the typical git work style will do many small commits to compose the result -- the user needs to check the result but not all steps.

The fourth one is a credentials vault which hides sensitive login details or access keys from the agent (\safeguard{CREDENTIAL VAULT SAFEGUARD}). This is a safeguard against a malicious agent or an agent affected by prompt injections or similar which would exfiltrate those secrets to any external third party.
First of all, this should protect the access key to the LLM service used for the agent and the login information (usually private ssh key) to the git forge. Even though the local clients running the agents need this access, the credentials need to be hidden from the agent. This can be achieved by intercepting the agent's communication and replacing the secrets.

The firewall safeguard and the git gateway safeguard are forms of selective control. The user does not need to supervise every action but can still allow or disallow certain actions.

\subsection{Safeguards vs. Risks}

\begin{figure}[t]
\centering
\includegraphics[width=12cm]{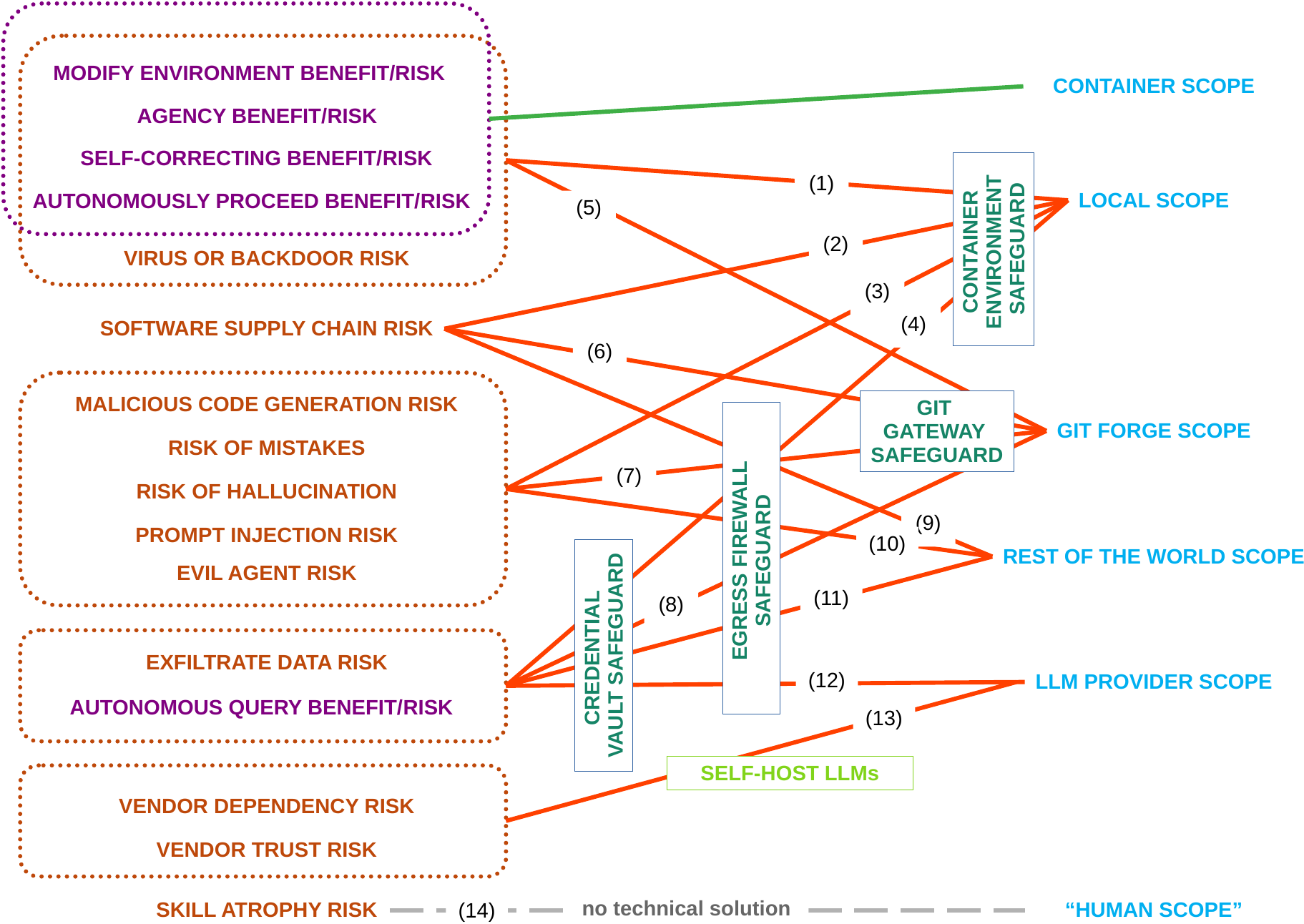}
\caption{Overview about the risks (left) that apply to the five scopes (right) and which of Terok's primary safeguards address them.}
\label{fig:risks_safeguards}
\end{figure}

The overview in \figurename~\ref{fig:risks_safeguards} shows how the safeguards address different groups of risks and the following text details the 12 applicable combinations plus 2 which are out of scope of our solution.

\noindent
(1) Protect the \scope{LOCAL SCOPE} against the agent through containerization: If the agent would modify its environment or try to take over the host like a conventional virus or backdoor would, then the user's environment or the host system outside of the container cannot be affected by the agent installing software, changing the configuration, etc.

\noindent
(2) Protect \scope{LOCAL SCOPE} against the \risk{SOFTWARE SUPPLY CHAIN RISK} through containerization: If any malicious software is installed inside the container as a dependency it cannot affect the host outside the container.

\noindent
(3) Protect the \scope{LOCAL SCOPE} against the agent through containerization: Regardless if it behaves maliciously or simply makes mistakes, the user's environment or the host system outside of the container cannot be affected by the agent's actions. The agent is free to run anything it wants.

\noindent
(4) Protect the \scope{LOCAL SCOPE} against exfiltration of data or credentials by containerization and by the egress firewall. First, the agent cannot access any information outside the container. And second, it cannot open connections to remote hosts unless authorized in the firewall by the user.

\noindent
(5-7) Protect the \scope{GIT FORGE SCOPE} against attempts by the agent to insert backdoors into the git repository (5), or insert dependencies to malicious software packages (6), or adding actual malicious code into the project itself (7).
This works through the git gateway where the user should check what is forwarded to the primary git repository.
Admittedly, that puts some burden on the user again and it is certainly not trivial to check. Especially if the agent tries it covertly. The advantage is that the user can do it in larger intermediate steps and does not need to check every step.

\noindent
(8) Protect the \scope{GIT FORGE SCOPE} from exfiltration of data. First through the git gateway, so that a branch which is not in the git gateway repo cannot be read by the agent. And second through the firewall again preventing unauthorized outgoing connections.

\noindent
(9-10) Protect the \scope{REST OF THE WORLD SCOPE} or actually users of the resulting software from malicious code inserted by the agent (9) or generated by the agent (10) in the same way as (5-7) through enforcing reviews by the user.

\noindent
(11) Protect authorized external destinations which the agent is allowed to communicate with. If the agent is supposed to work with sensitive data from there, then the firewall will prevent it from exfiltrating the data to unauthorized third parties.

\noindent
(12) This is a redundant safeguard for the API access key to the LLM (the \scope{LLM PROVIDER SCOPE}). It cannot be read by the agent due to the credential vault and even if it could, the firewall would prevent it from being sent to an unauthorized server. Even though this may seem like an overspecific issue, there are reports about LLM access keys as targets of software supply chain attacks, see the footnote about the LiteLLM attack above.

In addition, there are two aspects that cannot be solved through this concept:

\noindent
(13) When using LLM providers one has to trust them and will develop a dependency to their service. The LLM provider can see code as well as data and could send malicious commands to the agent. Furthermore, it could increase costs or terminate the service. There are no checks that would safeguard against this risk. Instead, one should think about a fallback which can be another provider or running self-hosted LLMs or at least maintain the ability to self-host LLMs. This is no “rocket science” but still comes with some costs and effort. Terok supports this directly through its provider-independent client configuration, including self-hosted and national-academic services such as Blablador (Helmholtz) and KISSKI (University of Göttingen) in Germany.

\noindent
(14) Finally, there is the risk that humans become dependent on AI assistance for coding and forget some skills when they no longer do it on a regular basis. This is also known as “skill atrophy”. This is certainly reversible because we did learn it in the past. It may not be a quick process, though. There is no technical solution against this and organizational solutions are outside of our scope here. On the bright side, this shows that Agentic AI for coding has the potential to free humans from a considerable amount of work and thus allow human developers to focus on other, maybe more interesting tasks.

\subsection{Implementation in Terok}

The Terok environment \url{https://github.com/terok-ai/terok} is our implementation of the presented protection concepts. It is an Open Source software environment that provides all described safeguards and adds a number of convenience features on top.

In IT security contexts, convenience falls into the wider security goal of “availability”, a goal naturally rivaling the other more immediate security goals fulfilled by the described safeguards. Especially in an end user application, finding the right compromise between convenience and security is a crucial design process for providing a responsible and secure system that will be accepted as a replacement for insecure systems. 
Therefore we paid special attention to convenience features and outline them here as well.

\subsubsection*{Implementation Overview}

\begin{figure}[t]
    \centering
    \includegraphics[width=0.8\linewidth]{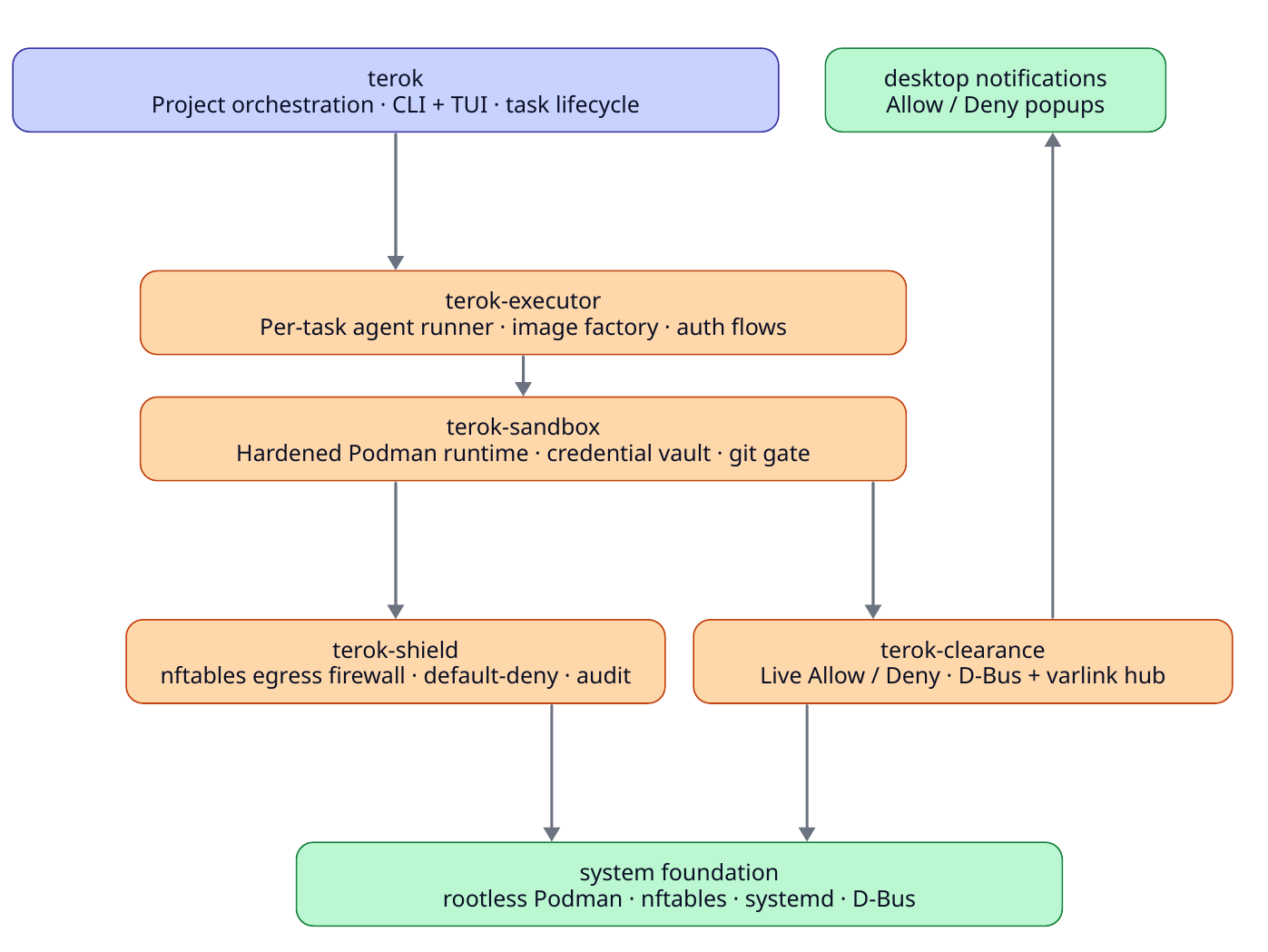}
    \caption{Overview of Terok's software components and how they interact}
    \label{fig:architecture}
\end{figure}

The four safeguards are implemented in the following way with the help of the named standard software components. An overview is shown in \figurename~\ref{fig:architecture}.

\noindent
\safeguard{CONTAINER ENVIRONMENT SAFEGUARD}: Terok runs every task (see below) in a separate rootless, hardened \software{Podman} container. For stronger isolation, the same task can instead run in a micro-VM via Podman's \software{krun} runtime.
This mode is offered as an experimental option while we close minor feature-parity gaps. 
Both modes use the same Podman container and configuration; only the container runtime differs, 
so switching runtimes needs no other architectural changes.

\noindent
\safeguard{EGRESS FIREWALL SAFEGUARD}: Terok uses the Terok-Shield component as a firewall for all outbound connections. It manages per-task allow-lists and can use either DNS names or IP addresses for approval and relies on \software{nftables}, \software{dnsmasq}, and \software{dig}. It can start with a curated allow-list (pre-allowed) and will present interactive allow/deny desktop notifications to the user. In addition, it keeps a detailed audit log.

\noindent
\safeguard{CREDENTIAL VAULT SAFEGUARD}: The credential vault will keep access keys, API access tokens, and other sensitive information secure and outside of the containers, where neither agent/LLM nor scripts or software executed by the agent could access it. When the secrets are used, they are inserted by proxying the communication from inside the container to an outside server and injecting the real secrets. For storing the secrets Terok relies on \software{systemd TPM2} which will store them in an encrypted way on the host using the host's trusted platform module (TPM) if available.

\noindent
\safeguard{GIT GATEWAY SAFEGUARD}: Terok provides a separate \software{git} repository for every task. It will be created and initialized from the primary git repository by Terok before the container is started using credentials provided by the user beforehand and kept in the credential vault. The agent/LLM from inside the container can pull and push from/to the gateway repository where the generated code is in quarantine. The user can access both, the gateway and the primary repository and can decide which commits or files to forward. It is also under the user's control when to pull new versions from the primary repository so that the agent can pull them from the gateway repository in turn.

\subsubsection{Convenience Add-ons}

The four main safeguards are integrated in a command line tool \texttt{terok} and the text user interface (TUI) \texttt{terok-tui} or a browser interface (web GUI). The installation is basically a single \texttt{pip} or \texttt{pipx} command followed by \texttt{terok setup}.\footnote{\url{https://github.com/terok-ai/terok\#quick-start}}

With \texttt{terok auth <provider>} one can configure the LLM API access keys globally, that means for all following projects and tasks. Then \texttt{terok project wizard} or the TUI can be used to create projects. For this the user needs to provide a name, a primary git repository (optional), and which container base image to start from (Terok currently supports "Fedora 44", "NVIDIA CUDA (GPU)", "Podman (Fedora-based)" and "Ubuntu 24.04"). If connecting to an upstream git repository, the wizard will generate a separate SSH key and ask the user to register the public key in the upstream GitHub or GitLab web UI.

\begin{figure}[t]
    \centering
    \includegraphics[width=0.95\textwidth]{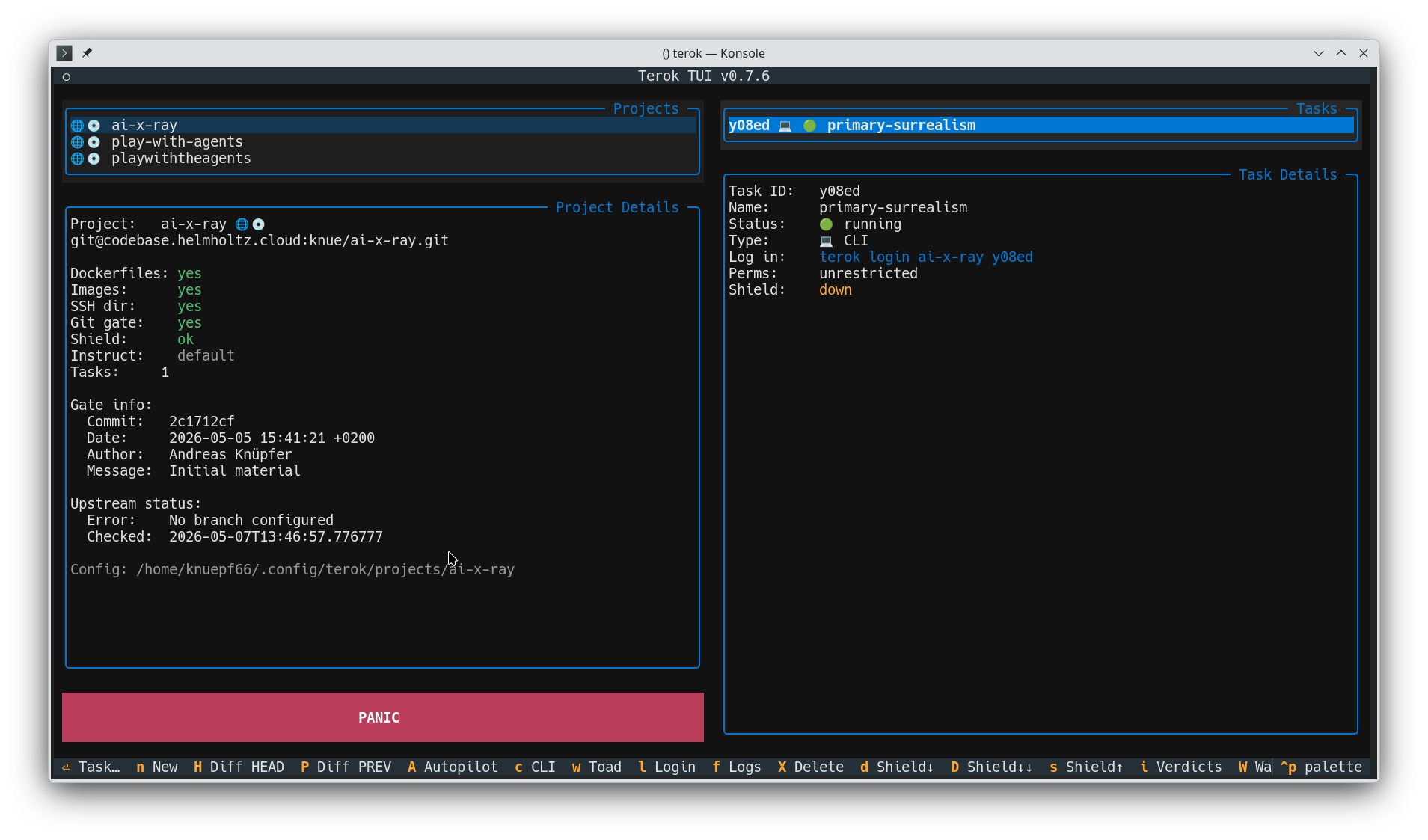}
    \caption{Screenshot of Terok's text user interface (TUI) for managing projects and tasks. The projects panel (left) shows multiple projects and the details of one of them. The tasks panel (right) shows one task belonging to the selected project.}
    \label{fig:tui}
\end{figure}

From a single project, one can create and manage one or multiple tasks, each of them being an independent session that works (by default) isolated from the others.
Each task gets its own container instance and git gateway. Essentially, this is the playground for one agent.
See \figurename~\ref{fig:tui} for an impression of the Terok TUI with the projects and tasks management view. 

\begin{figure}[t]
    \centering
    \includegraphics[width=0.95\textwidth]{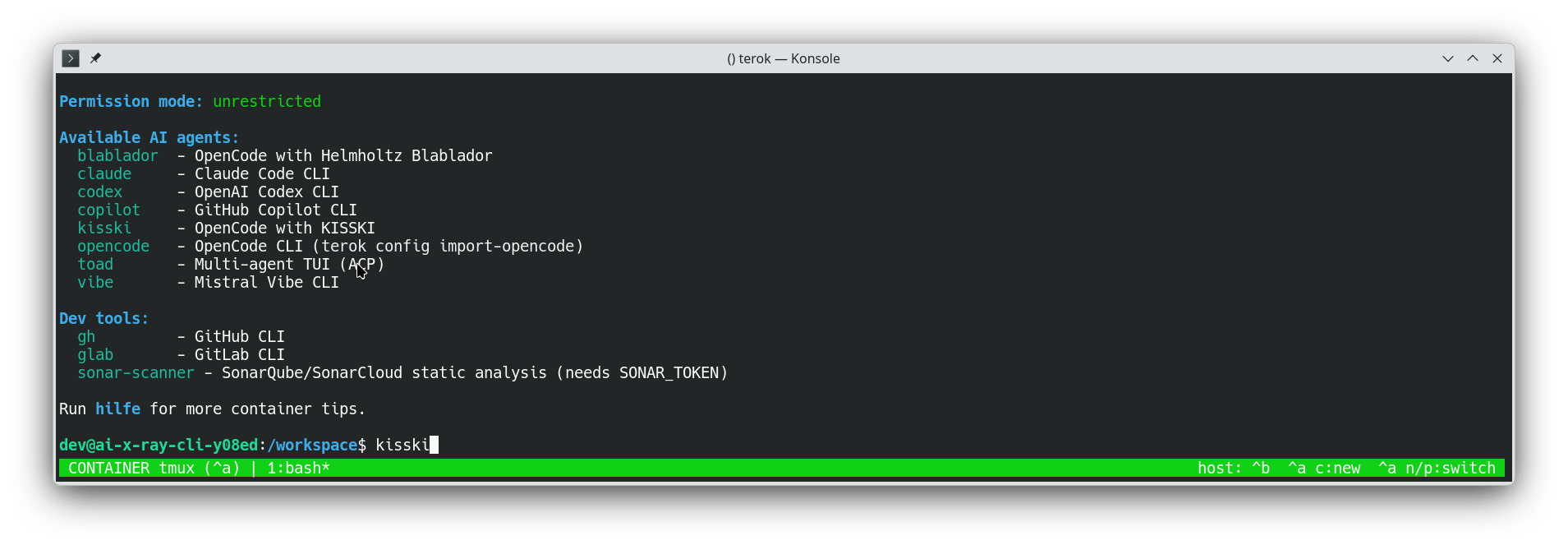}
    \caption{Screenshot of Terok's text user interface (TUI) with the shortcut commands to call one AI agent from a selection of providers.}
    \label{fig:commands}
\end{figure}

Inside each task in the separate container instance, Terok uses existing tools. The default is simply a shell session inside \software{tmux}, a terminal multiplexer. The other options are LLM clients like \software{opencode} (an Open Source software supporting different LLM providers), \software{claude} (the Anthropic client), \software{codex} (the OpenAI client), \software{copilot} (the GitHub Copilot client), \software{vibe} (the Mistral client) or others. In addition there are shortcut commands which use \software{opencode} with specific providers like OpenRouter, Blablador (from the Helmholtz association of the German national labs), or KISSKI (from University of Göttingen, Germany). All of them are ready-made to be used with the API access tokens that were configured by Terok outside of the container. See the screenshot in \figurename~\ref{fig:commands} for an impression. 
The diversity of providers and agents/LLMs supported should simplify and encourage experimenting with different models for certain tasks to allow an assessment of quality and ability as much as possible.

The user interface in each of the clients is the native and unmodified client software. This means that new or improved versions of the vendor clients can be installed by Terok automatically without any delay.

\section{Summary and Outlook}

The Terok environment sandboxes AI Agents, supports vendor LLMs and self-hosted LLMs, and adds convenience for managing projects and sessions. Now we can use Agentic AI as a coding tool in a responsible manner for exploring its full potential and for actual coding.

Is this the once-and-for-all solution? Probably not because it is a very fast-moving field. We will continue to work on it, but it is a substantial step towards responsible usage already.


As we only started exploring the potentials and consequences of agentic AI software development, we will continue to watch innovations in this field and adapt the concept and the Terok project. 
Promising ongoing steps are more sophisticated interactions between agents, from planning agent vs. sub-agents to even more diverse roles. This may bring implications for the security architecture.
With respect to usability we plan to extend the git gateway repositories further to elevate them to a more complete proxy for git forges. If they could funnel pull requests (PRs) and issues in both directions without opening security risks, they'd become more convenient for users and for agents.

We invite the interested community to try Terok from \url{https://github.com/terok-ai/terok}. We are happy about feedback, ideas, fixes, and opinions, both on the conceptual level and on the implementation level.

\subsection*{Statement on AI Support}

This manuscript has been written by the human authors who take full responsibility for the content. AI assistance  was used for researching related work (ScienceOS \url{https://www.scienceos.ai/}) and for spellchecking.

\printbibliography

\end{document}